%% file: main.tex
\documentclass[10pt,twocolumn,letterpaper]{article}

\usepackage{wacv}
\usepackage{times}
\usepackage{epsfig}
\usepackage{graphicx}
\usepackage{amsmath}
\usepackage{amssymb}
\usepackage{booktabs}
\usepackage{color}
\usepackage{soul}
\usepackage{xcolor}
\usepackage{url}
\usepackage{cite}
\usepackage{floatrow}
\usepackage[draft,inline,nomargin]{fixme}
\fxsetup{
    status=draft,
    author=,
    layout=inline,
    theme=color
}

\wacvfinalcopy % *** Uncomment this line for the final submission

\def\wacvPaperID{226} % *** Enter the wacv Paper ID here

\newcommand{\xhdr}[1]{\vspace{6pt} \noindent {\textbf{#1} }}
\newcommand{\ra}[1]{\renewcommand{\arraystretch}{#1}}

\ifwacvfinal\fi
\begin{document}

%%%%%%%%% TITLE
\title{Multi-Task Spatiotemporal Neural Networks \\ for Structured Surface Reconstruction}

% Authors at the same institution
%\author{First Author \hspace{2cm} Second Author \\
% Authors at different institutions
\author{Mingze Xu$^{1}$
    \hspace{0.3cm} Chenyou Fan$^{1}$
    \hspace{0.3cm} John D. Paden$^{2}$
    \hspace{0.3cm} Geoffrey C. Fox$^{1}$
    \hspace{0.3cm} David J. Crandall$^{1}$ \\
    $^{1}$Indiana University, Bloomington, IN \\
    $^{2}$University of Kansas, Lawrence, KS \\
    {\tt\small \{mx6, fan6, gcf, djcran\}@indiana.edu, paden@ku.edu}
}

\maketitle
\ifwacvfinal\thispagestyle{empty}\fi

%%%%%%%%% ABSTRACT
\begin{abstract}
    Deep learning methods have surpassed the performance of
    traditional techniques on a wide range of problems in computer
    vision, but nearly all of this work has studied consumer photos,
    where precisely correct output is often not critical. It is less
    clear how well these techniques may apply on structured prediction
    problems where fine-grained output with high precision is
    required, such as in scientific imaging domains. Here we consider
    the problem of segmenting echogram radar data collected from the
    polar ice sheets, which is challenging because segmentation
    boundaries are often very weak and there is a high degree of
    noise. We propose a multi-task spatiotemporal neural network that
    combines 3D ConvNets and Recurrent Neural Networks (RNNs) to
    estimate ice surface boundaries from sequences of tomographic
    radar images. We show that our model outperforms the
    state-of-the-art on this problem by (1) avoiding the need for
    hand-tuned parameters, (2) extracting multiple surfaces (ice-air
    and ice-bed) simultaneously, (3) requiring less non-visual
    metadata, and (4) being about 6 times faster.
\end{abstract}

%-------------------------------------------------------------------------
\input{introduction}
\input{related_work}

\input{methodology}
\input{experiments}
\input{conclusion}
\input{acknowledgements}

{\small
\bibliographystyle{ieee}
\bibliography{egbib}
}

\end{document}

%% file: introduction.tex
\vspace{-5pt}
\section{Introduction}

Three-dimensional imaging is widely used in scientific research
domains (e.g., biology, geology, medicine, and astronomy) to 
characterize the structure of objects and how they change over time.
Although the exact techniques differ depending on the problem and
materials involved, the common idea is that electromagnetic waves (e.g., X-ray, radar, etc.) 
are sent into an object, and signal returns in the form of sequences
of tomographic images are then analyzed to estimate the object's 3D structure.
However, analysis of these image sequences can be difficult even
for humans, since they are often noisy and require
integrating evidence from multiple sources simultaneously.

%% large-scales is difficult nearly impossible even for human experts,
%% since they are full of noise and usually require a pixel-level
%% analysis is impossible for eyes.  Computer Vision methods are then
%% applied for automated labeling and analyzing the region of interest as
%% both efficient and effective ways. As a particular example,
%% glaciologists are interested in analyzing large-scale polar ice sheet
%% data,to understand and predict the effect of melting glacier. A ground
%% penetrating radar system is use to capture the material (e.g. air, ice
%% and terrain) boundaries information by flying over the ice sheet
%% ground. The returned raw data is a group of 2D radar images that
%% capture the cross-sectional images of underground.

% Overall architecture
\begin{figure}
    \begin{center}
        \includegraphics[width=1\columnwidth]{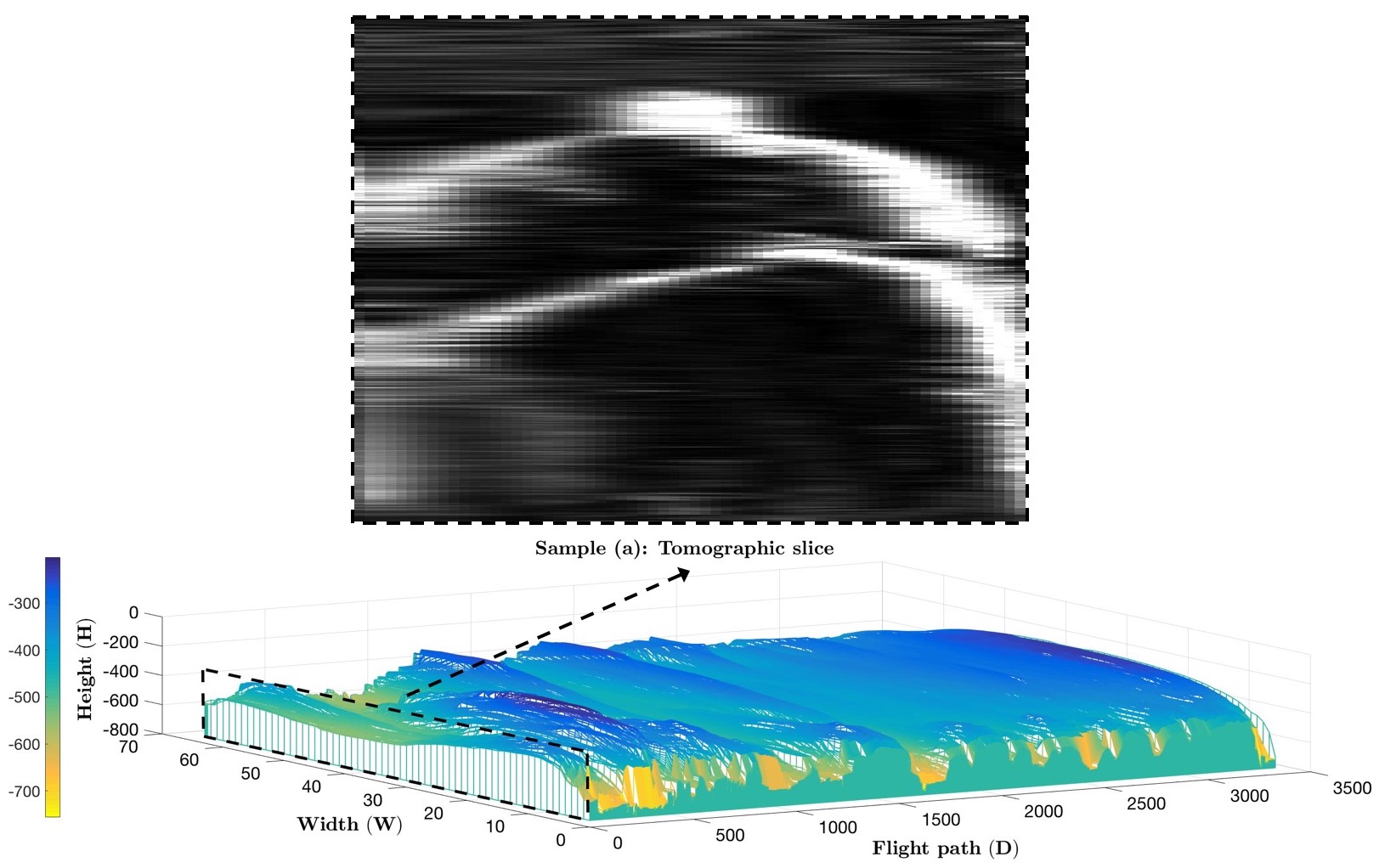}
    \end{center}
    \vspace{-5pt}
    \caption{
        Illustration of our task. A ground-penetrating radar system flies over
        a polar ice sheet, yielding a
        sequence of 2D tomographic slices (e.g. Sample (a) with the black dashed
        bounding box). Each slice captures a vertical cross-section of the ice,
        where two material boundaries (the ice-air and ice-bed layer) are visible
        as bright curves in the radar echogram. Given such a sequence of tomographic
        slices, our goal is to reconstruct the 3D surfaces for each material
        boundary (e.g. a sample ice-bed surface~\cite{icesurface2017icip} is shown in the figure).
    }
    \label{fig:teaser}
\end{figure}

% % Overall architecture
% \begin{figure}
%     \begin{center}
%         \includegraphics[width=1\columnwidth]{figures/teaser.jpg}
%     \end{center}
%     \caption{A sample tomographic image taken by a ground-penetrating radar system over a polar ice sheet.
%       The image captures a vertical cross-section of the ice, where
%     two material boundaries (the ice-air and ice-bed layer) are visible as bright curves in the radar echogram.}
%     \vspace{-10pt}
%     \label{fig:teaser}
% \end{figure}

As a particular example, an important part of modeling and forecasting
the effects of global climate change is to understand polar
ice. Hidden beneath the ice of the poles 
is a rich and complex structure:  the ice consists of
multiple layers that have accumulated over many
thousands of years, and the base is bedrock that has a complicated
topography just like any other place on Earth (with mountains,
valleys, and other features).  Moreover, the ice sheets move over time, and their movement
is determined by a variety of factors, including temperature changes,
flows underneath the surface, and the topography of the bedrock below
and nearby. Accurately estimating all of this rich structure is
crucial for understanding how ice will change over time, which in turn
is important for predicting the effects of melting ice associated with
climate change.

Glaciologists traditionally had to drill ice cores to probe the
subsurface structure of polar ice, but advances in ground-penetrating
radar technology have revolutionized this data collection process.
But while these radar observations can now be collected over very large areas,
actually analyzing the radar data to determine the structure of
subsurface ice is typically done by hand~\cite{macgregor}. This is
because the radar echograms produced by the data collection process
are very noisy: thermal radiation, electromagnetic interference, complex ice
composition, and signal attenuation in ice, etc.\ affect radar signal returns
in complex ways.
Relying on humans to interpret data not only limits the rate at which
datasets can be processed, but also limits the type of analysis that
can be performed: while a human expert can readily mark ice sheet
boundaries in a single 2D radar echogram, doing this simultaneously
over thousands of echograms to produce a 3D model of an ice bed,
for example, is simply not feasible.

While several recent papers have proposed automated techniques for
segmenting layer boundaries in ice~\cite{freeman2010automated,
  ilisei2012technique, ferro2013automatic,
  mitchell2013semi,crandall2012layer, lee2014estimating,
  carrer2017automatic, icesurface2017icip, panton}, none have
approached the accuracy of even an undergraduate
student annotator~\cite{macgregor}, much less an expert. However, these
  techniques have all relied on traditional image processing and
  computer vision techniques, like edge detection, pixel template
  models, active contour models, etc. Most of these techniques also
  rely on numerous parameters and thresholds that must be tuned by
  hand. Some recent work reduces the number of free parameters through
  graphical models that explicitly model noise and
  uncertainty~\cite{icesurface2017icip, lee2014estimating,
    crandall2012layer, panton} but still rely on simple
  features.

In this paper, we apply deep networks to the problem of ice
boundary reconstruction in polar radar data. Deep networks have become the
\textit{de facto} standard technique across a wide range of vision
tasks, including pixel labeling problems.  The majority of these
successes have been on consumer-style images, where there is
substantial tolerance for incorrect predictions. 
In contrast, for problems involving scientific datasets
like ice layer finding, there is typically only one ``correct''
answer, and it is important that the algorithm's output be as accurate as
possible.

Here we propose a technique for combining 3D convolutions and
Recurrent Neural Networks (RNNs) to perform segmentation in 3D,
borrowing techniques usually used for video analysis to instead
characterize sequences of tomographic slice images.  In particular, since
small pixel value changes only affect a few adjacent images, we
apply 3D convolutional neural networks to efficiently capture
cross-slice features. We extract these spatial
and temporal features for small neighborhoods of slices, and then apply
an RNN for detailed structure labeling across the entire 2D image.
Finally,
layers from multiple images are concatenated to generate a 3D surface estimate.
We test our model on extracting 3D ice subsurfaces from sequences of radar
tomographic images, and achieve the state-of-the-art results in both
accuracy and speed.

%% file: related_work.tex
\section{Related Work}

A number of methods have been developed for detecting layers or
surfaces of material boundaries from sequential noisy radar images.
For example, in echograms from Mars, Freeman et
al.~\cite{freeman2010automated} find layer boundaries by applying
band-pass filters and thresholds to find linear subsurface structures,
while Ferro and Bruzzone \cite{ferro2011novel} identify subterranean
features using iterative region-growing. Crandall et
al.~\cite{crandall2012layer} detect the ice-air and ice-bed layers
in individual radar echograms by combining a pre-trained template
model and a smoothness prior in a probabilistic graphical model. In
order to achieve more accurate and efficient results, Lee et
al.~\cite{lee2014estimating} utilize Gibbs sampling from a joint
distribution over all candidate layers, while Carrer and
Bruzzone~\cite{carrer2017automatic} reduce the computational
complexity with a divide-and-conquer strategy.  Xu et
al.~\cite{icesurface2017icip} extend the work to the 3D domain to
reconstruct 3D subsurfaces using a Markov Random Field (MRF).

In contrast, we are not aware of any work that has studied this
application using deep neural networks. In the case of segmenting
single radar echograms, perhaps the closest analogue 
is segmentation in consumer images~\cite{tseng2017joint}. Most of
this work differs from the segmentation problem we consider here,
however, because our data is much noisier, our ``objects'' are much
harder to characterize (e.g., two layers of ice look virtually
identical except for some subtle changes in texture or intensity), our
labeling problem has greater structure, and our tolerance for
errors in the output is lower.

% Overall architecture
\begin{figure*}
    \begin{center}
        \includegraphics[width=17.5cm]{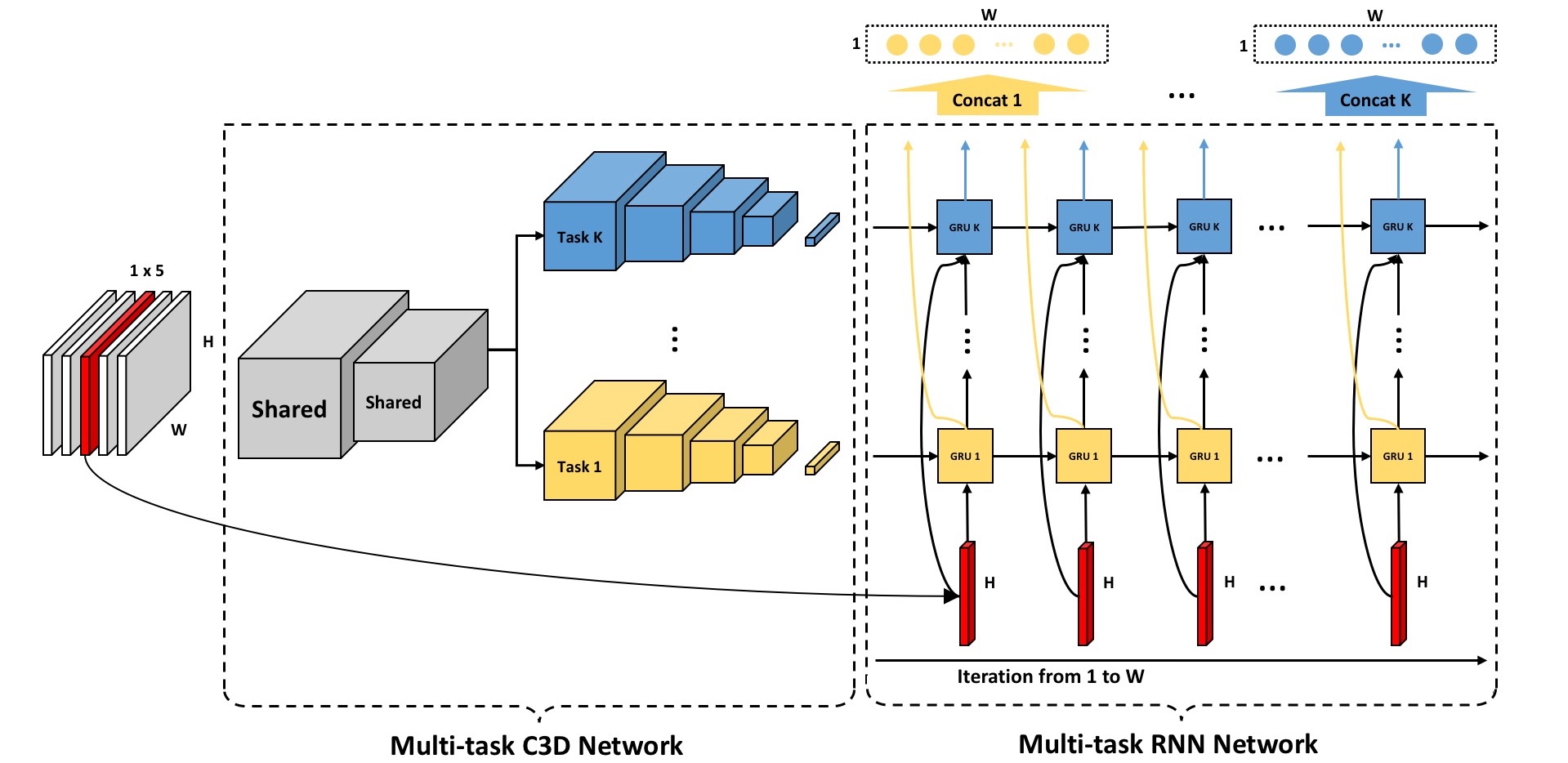}
    \end{center}
    \vspace{-10pt}
    \caption{Architecture of our model for predicting multiple ice layers in tomographic images. We extract and reconstruct structured 3D surfaces from sequential data by combining C3D and RNN networks. A C3D network serves as a robust feature extractor to capture both local within-slice and between-slice features in 3D space, and an RNN serves to capture longer-range structure both within individual images and across the entire sequence.}
\label{fig:overall}
\end{figure*}

For segmenting 3D regions, perhaps the closest related work is in deep
networks for video analysis, where the frames of video can be viewed
as similar to our tomographic slices.  Papers that apply deep networks
to video applications focus on efficient ways to combine spatial and
temporal information, and can be roughly categorized into
three classes: (1) combining both RGB frames for spatial features and optical
flow images for temporal features in two-stream
networks~\cite{simonyan2014very}, (2) explicitly learning 3D spatiotemporal
filters on image spaces through techniques such as C3D~\cite{tran2015learning}, and (3)
various combinations of both \cite{carreira2017quo}. In order to obtain
video representations from per-frame or per-video-segment features, it
is a common practice to apply temporal pooling to
abstract into fixed-length per-video
features~\cite{karpathy2014large,simonyan2014very}. These approaches
achieve significantly better classification accuracy on video
classification compared to traditional approaches using
hand-crafted features.

Recurrent Neural Networks (RNNs) and the specific version we consider
here -- Gated Recurrent Units (GRUs) -- have been proposed for
learning sequential data, such as natural language
sentences~\cite{elman1990finding, graves2013generating}, programming
language syntax~\cite{karpathy2015visualizing}, and video
frames~\cite{yue2015beyond}.  A popular application of RNNs
recently~\cite{vinyals2014show, karpathy2014deep} is to generate image
captions in combination with CNNs. In this case, CNNs are used to
recognize image content while RNNs are used as language models to
generate new sentences. Video can also be thought of as sequential data, since
adjacent frames share similar content while differences reveal
motion and other changes over time. A large variety of
studies~\cite{yue2015beyond,donahue2015long, piergiovanni2017learning}
share the common idea of applying RNNs on deep features for each
video frame and pooling or summing over them to create a video descriptor.
 Other successful applications of RNNs to
interesting vision and natural language tasks
include recognizing multiple objects by making guided
glimpses in different parts of images~\cite{ba2014multiple}, answering
visual
questions~\cite{andreas2016learning,weston2015towards,kumar2016ask},
generating new images with
variations~\cite{gregor2015draw,yu2017seqgan}, reading
lips~\cite{chung2016lip}, etc.

We build on this existing work but apply to the novel domain of  extracting and reconstructing structured 3D surfaces
    from sequential data by combining C3D and RNN networks. In
    particular, we use the C3D network as a robust feature extractor
    to capture local-scale within-slice and between-slice features in 3D
    space, and use the RNN to capture longer-range structure both within single slices and across the entire image sequence.

%% file: methodology.tex
\section{Technical Approach}

Three-dimensional imaging typically involves sending electromagnetic radiation (e.g., radar, X-ray, etc.)
into a material and collecting a sequence of cross-sectional tomographic slices
$I = \{I_1, I_2, \cdots, I_D\}$ that characterize returned signals along the
path. 
Each slice $I_d$ is a 2D tomographic image of size $H \times W$ pixels.
In the particular case of ice segmentation, we are interested in locating $K$ \textit{layer surface boundaries}
between different materials.
Our output surfaces are highly structured,
since there should be exactly $K$ surface pixels within any column of a given
tomographic image. 
We thus need to estimate the layer boundaries in each individual
slice, while incorporating evidence from all slices jointly in order to overcome noise and resolve ambiguities. Layer
boundaries within each slice can then be concatenated across slices to produce a 3D surface.

In this section, we describe the two important components of our network framework:
our
multi-task 3D Convolutional (C3D) Network that captures within-slice features as well as evidence
from nearby slices, and
our Recurrent Neural Network (RNN) which incorporates longer-range cross-slice constraints.
The overall architecture is shown
in Figure~\ref{fig:overall}.

\begin{figure*}
    \begin{center}
        \includegraphics[width=17.5cm]{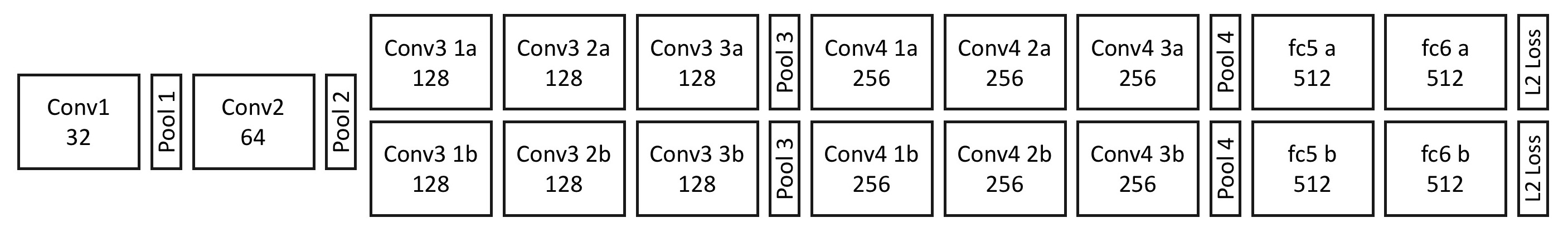}
    \end{center}
    \vspace{-10pt}
    \caption{Illustration of our C3D architecture in a special case of two layers ($K = 2$).
    All 3D convolution kernels are $3 \times 5 \times 3$ with stride $1$ in
    each dimension and the 3D pooling kernels are $1 \times 2 \times 1$ with
    stride $2$ in the height dimension of each image.
}
    \label{fig:c3d}
\end{figure*}

\subsection{A Multi-task C3D Architecture}
Traditional convolutional networks for tasks like
object classification and recognition 
lack the ability to model spatiotemporal features in 3D space.
More importantly, their use of max or average pooling operations makes it
impractical to preserve temporal information within the sequential inputs.
To address these problems, we use C3D networks to capture local spatiotemporal
features in our sequence of input images.
%Compared to the C2D network, 3D
%convolution and 3D pooling operations have more advantages in modeling sequential
%dependencies between consecutive images. 
C3D has typically been used for video, but our dataset has very similar characteristics: 
we have a sequence of tomographic slices
taken in consecutive (discrete) positions along the path of a
penetrating wave source (a moving airplane, in the case of our ice application). Physical constraints on 
layer boundaries (e.g., that they should be continuous and generally smooth) mean that integrating information across adjacent
images improves accuracy, especially when data within any give slice is particularly noisy or weak.
%\hl{To be clear, rather than using one single image as input in 2D
%convolution, we apply 3D convolution on a small sequence of images to detect
%layer boundaries for the middle image.}

Figure~\ref{fig:c3d} illustrates details of our C3D architecture, which is based on Tran et al.~\cite{tran2015learning} but 
with several important modifications.
Since the features of these structured layers in
tomographic images are typically less complicated than consumer photos, we use a simpler
network architecture, as follows.
For the input, our model takes $L$ consecutive images,
where $L$ is a small odd number; we have tried
$L = 1, 3, 5, \cdots, 11$, and choose $5$ as the best empirical balance between
running time and accuracy. Then, we use two shared convolutional layers,
each of which is followed by rectifier (ReLU) units and max pooling operations,
to extract low-level features for all layers. The key idea is that different
kinds of layer boundaries usually share similar detailed patterns, although they have
different high-level features, e.g., shapes. Inspired by the template model
used in Crandall et al.~\cite{crandall2012layer} and Xu  et al.~\cite{icesurface2017icip}, our model uses
rectangular convolutional filters with a size of $3\times5\times3$, since the important features lie along
the vertical dimension.
Afterwards, the framework is divided into $K$ branches, each with
6 convolutional layers for modeling features specific to each type of ice layer boundary. The filter
size is the same as with the shared layers. Two fully-connected layers are appended to
the network for each ice layer, where the $k$-th ice layer has $W$ outputs $S^k_d =
\{s^k_{d,1}, s^k_{d,2}, \cdots, s^k_{d,W}\}$, each corresponding to a column of the
tomographic slice $I_d$, 
representing the row coordinate of the $k$-th ice layer boundary within that column.
All training images have been labeled with ground
truth vectors, $G^k_d = \{g^k_{d,1}, g^k_{d,2}, \cdots, g^k_{d,W}\}$ to
indicate the correct position of these output layers in each image.

\begin{figure}
    \begin{center}
        \includegraphics[width=8.5cm]{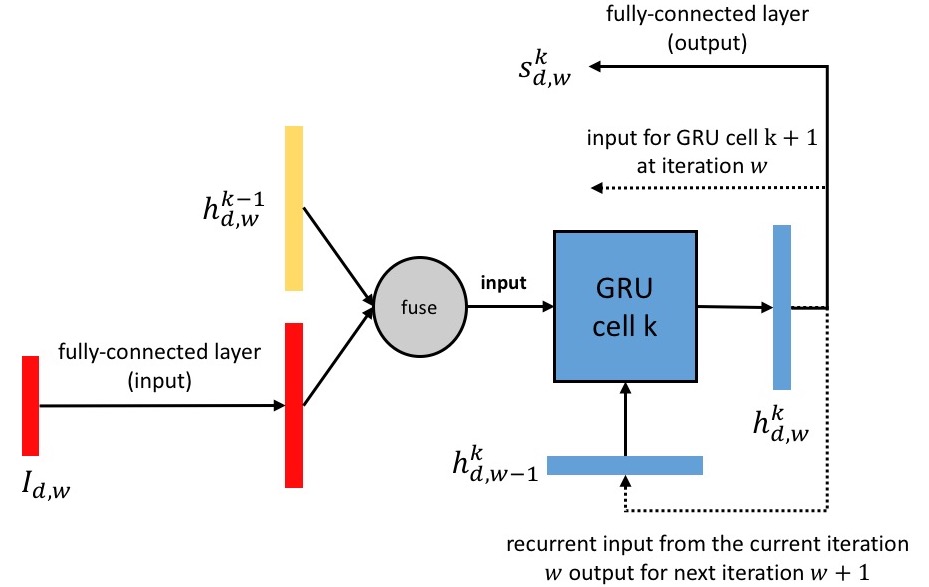}
    \end{center}
    \vspace{-12pt}
    \caption{
        Visualization of the $k$-th GRU at iteration $w$.
}
    \label{fig:gru}
\end{figure}

We train the C3D network using the L2 Euclidean loss $L_{elu}$ to encourage the
model to predict correct labelings according to human-labeled ground truth,
\begin{equation}
    L_{elu} = \frac{1}{2} \sum_{k=1}^K \sum_{w=1}^W (s^k_{d,w} - g^k_{d,w})^2.
\end{equation}
We note that this formulation differs from most semantic
and instance segmentation work which typically uses Softmax and Cross-entropy as the
target function. This is because we are not assigning each pixel to a
categorical label (e.g., cat, dog, etc.), but instead assigning each column of
the image with a row index. Since these labels are ordinal and continuous,
it makes sense to directly compare them and minimize a Euclidean loss.

% Overall architecture
\begin{figure*}
    \begin{center}
  \begin{tabular}{@{}p{12pt}@{}l}
\rotatebox{90}{\textbf{\hspace{35pt}Ours \hspace{50pt} Human-labeled}} &
        \includegraphics[width=17cm]{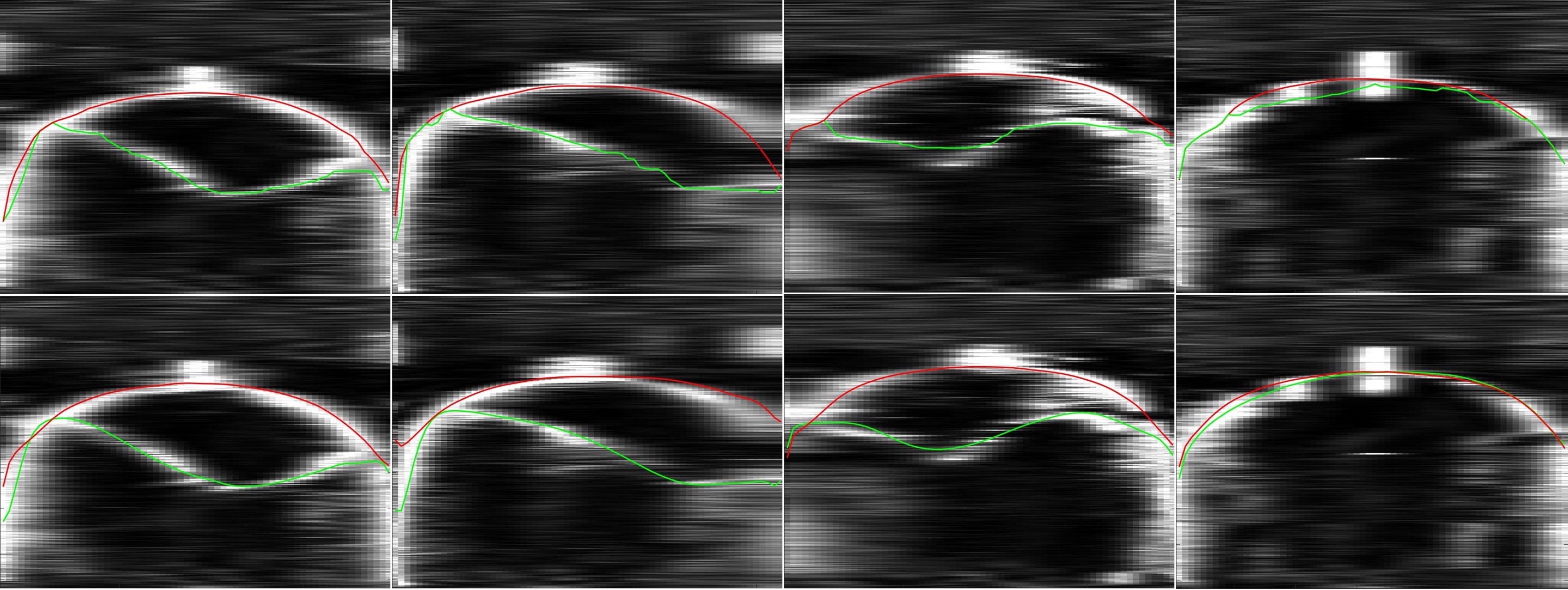}\\
\end{tabular}
    \end{center}
\vspace{-10pt}
    \caption{
        Visualization of sample tomographic images with height $H$ and width $W$.
        The first row shows the ice-air (red) and ice-bed (green) layers labeled
        by human annotator, while the second row shows the predicted layers by
        our model. In general, our predictions not only capture the precise location
        of each ice layer, but are also smoother than human labels.
}
    \label{fig:slices}
\end{figure*}

\subsection{A Multi-task RNN Architecture}
\label{sec:rnn}
The C3D networks discussed above model features both in the temporal
and spatial dimensions, but only in very small neighborhoods. For
example, they can model the fact that adjacent pixels within the same
layer should have similar grayscale value, but not that the layer
boundaries themselves (which are usually separated by dozens of pixels
at least) are often roughly parallel to one another.  Similarly, C3D 
models some cross-slice constraints but only in a few slices in
either direction.  We thus also include an RNN that incorporates
longer-range cross-slice evidence.  Because of the limited training
data, we use Gated Recurrent Units (GRUs)~\cite{chung2015gated} since
they have fewer learnable parameters than other popular networks
like Long Short-Term Memories (LSTMs)~\cite{hochreiter1997long}.

\xhdr{GRU Training and Testing.}
The multi-task GRU framework is shown in Figure~\ref{fig:overall}. Our model for each individual slice consists of
$K$ GRU cells, each responsible for predicting the $k$-th layer in each image.
Each GRU cell takes a tomographic slice $I_d$ and the output of the previous GRU layer as inputs,
and produces $W$ real value numbers indicating the predicted positions of the layer within each column of the image.
Each GRU also takes as input the output from the GRU corresponding to the same ice layer in the \textit{previous} slice,
since these layer boundaries should be continuous and roughly smooth.
In previous work~\cite{crandall2012layer, lee2014estimating, icesurface2017icip}, this prior knowledge was explicitly enforced by
pairwise interaction potentials, which were manually tuned by human experts.
Here we train RNNs to be able to model more general  relationships in a fully learnable way.
%capture the internal relationship between adjacent
%pixel positions as well as relative positions between different layers by ``reading''
%the context of the layer configuration.

We split each tomographic input image
$I_{d}$ into separate column vectors
$I_{d,w}$, $w = 1, 2, \cdots, W$, each with width $1$ and height $H$.
Each column vector is projected to the length of the GRU hidden
state with a fully-connected layer. During training time, the $k$-th GRU cell is operated
for $W$ iterations, where each iteration $w$ predicts the $k$-th layer position
in image column $I_{d,w}$. Then in a given iteration $w$, the $k$-th GRU
takes the fused features (e.g., using sum or max fusion) of
the (resized) image column $I_{d,w}$ and the hidden state $h^{k-1}_{d,w}$ as the input.
It also receives the hidden states $h^{k}_{d, w-1}$ of itself in iteration $w-1$
as contextual information. More formally, the $k$-th GRU cell outputs a sequence
of hidden states $h^{k}_{d,1}, h^{k}_{d,2}, \cdots, h^{k}_{d,W}$ with iteration
$w = 1, 2, \cdots, W$, and each hidden state $h^{k}_{d,w}$ is followed by a fully-connected
layer to predict the actual layer position $s^{k}_{d,w}$ as shown in Figure~\ref{fig:gru}.
Since each GRU has the same operation for each 2D image $I_d$,
we drop $d$ subscript for simplicity, and compute,
% \hl{Could you maybe check this paragraph? I'm having trouble understanding it, and i wonder if
% some rewording would help.}
% At beginning, we shatter each entire tomographic image to columns $I_{d,w}$
% that has width $1$ but of original height $H$, and project each image column $I_{d,w}$
% into the same dimensional space with the GRU hidden layer by using a fully-connected layer.
% During training time, each GRU cell $k$ fuses $\mathcal{F}$ each image column $I_{d,w}$ and
% the hidden state $h^{k-1}_{d,w}$ from the ($k$-$1$)-th GRU as an input, as well
% as previous hidden state $h^{k}_{d,w-1}$ as contextual information to predict
% $k$-th layer position $s^k_{d,w}$ at this column $w$.
% Then, for the $k$-th GRU cell, it computes a
% sequence of hidden states $(h^k_{d,1}, h^k_{d,2}, \cdots, h^k_{d,w})$ and
% outputs a 2D layer $(s^k_{d,1}, s^k_{d,2}, \cdots, s^k_{d,w})$ by iterating the
% following recurrence for $w = 1, 2, \cdots, W$. Since the GRU has the same
% operations for each 2D image $I_d$, we drop $d$ subscribe for simplicity.
\begin{equation*}
\begin{split}
    &z_w = \textrm{sigmoid} (U_{iz} \mathcal{F}(I_{w},h^{k-1}_{w}) + U_{hz} h_{w-1} + b_{z}), \\
    &r_w = \textrm{sigmoid} (U_{iz} \mathcal{F}(I_{w},h^{k-1}_{w}) + U_{hz} h_{w-1} + b_{r}), \\
    &n_w = \textrm{tanh}    (U_{in} \mathcal{F}(I_{w},h^{k-1}_{w}) + U_{hn} (r_{w} \circ h_{w-1}) + b_{n}), \\
    &h_w = z_{w} \circ h_{w-1} + (1 - z_{w}) \circ n_{w}, \, \textrm{and} \\
    &s_w = U_{y} h_{w} + b_{y},
\end{split}
\end{equation*}
where $\circ$ is the Hadamard product, $z_{w}$, $r_{w}$, $n_{w}$, $h_{w}$, and
$s_{w}$ are the reset, input, new gate, hidden state, and output layer position at time $w$,
respectively. We use 512 neurons in the hidden layer of the GRU.
We train the GRU network with the same L2 Euclidean loss $L_{elu}$ as discussed
in the previous section.

% Echogram figure
\begin{figure*}[]
    \centerline{\includegraphics[width=17.5cm]{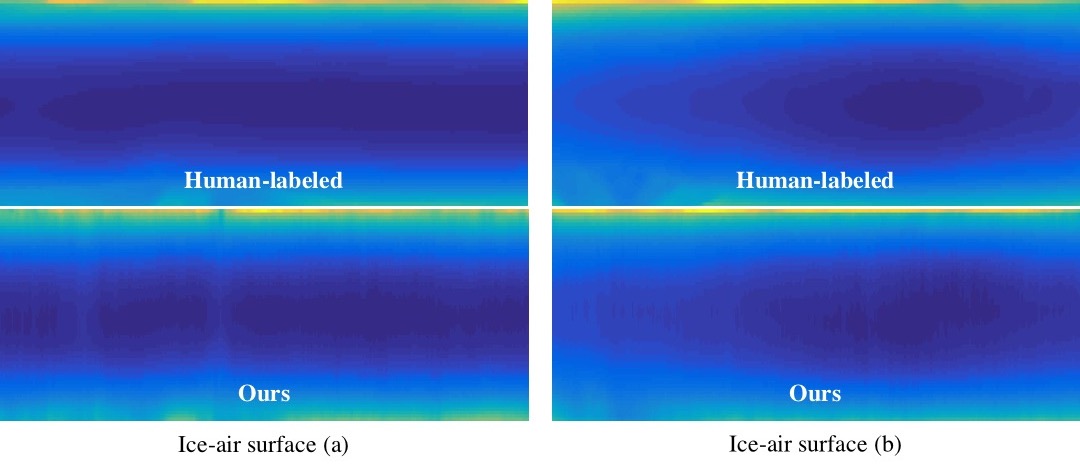}}
    \vspace{-8pt}
    \caption{
Results of the extracted ice-air surfaces based on
        about 330 tomographic images. 
 The x-axis corresponds to distance along
the flight path, the y axis is the width of the tomographic images ($W$),
and the color is the height dimension (max height is $H$), which also represents
the depth from the radar.}
    \label{fig:ice_air}
\end{figure*}

\subsection{Combination}
We combine our proposed C3D model and GRU model for efficiently encoding
spatiotemporal information into explicit structured layer predictions.
%The idea is that we use a neighborhood of tomographic images over time of
%current frame to provide high-level information such as trend and shape of ice
%layers in current frame; then we accurately identify positions of those layers
%by examining current frame in column-wise fashion. In practice, w
We use the C3D features $\textrm{C3D}^{k}_{\theta}(I_{d,k})$ (where $\textrm{C3D}^{k}_{\theta}$
denotes the features with model parameters $\theta$ for the
$k$-th ice layer) to initialize the $k$-th GRU's hidden state
$h_{1}$, as shown in Figure~\ref{fig:overall}. In the figure, $I_d$ is marked
in red; this is the frame currently under consideration, which is divided
into columns which are then provided to the GRU cells one at a time.
% shows the case when there are K
%layers. K branches of C3D networks provide global trend of these K
%layers; each of which is used to initialize hidden state of
%corresponding GRU cells. In the meantime, current frame (marked as
%red) is shattered to columns which are provided to GRU cells once a
%time to predict the pixel position which the layer falls on.

%% file: experiments.tex
\section{Experiments}

% Echogram figure
\begin{figure*}[]
    \centerline{\includegraphics[width=17.5cm]{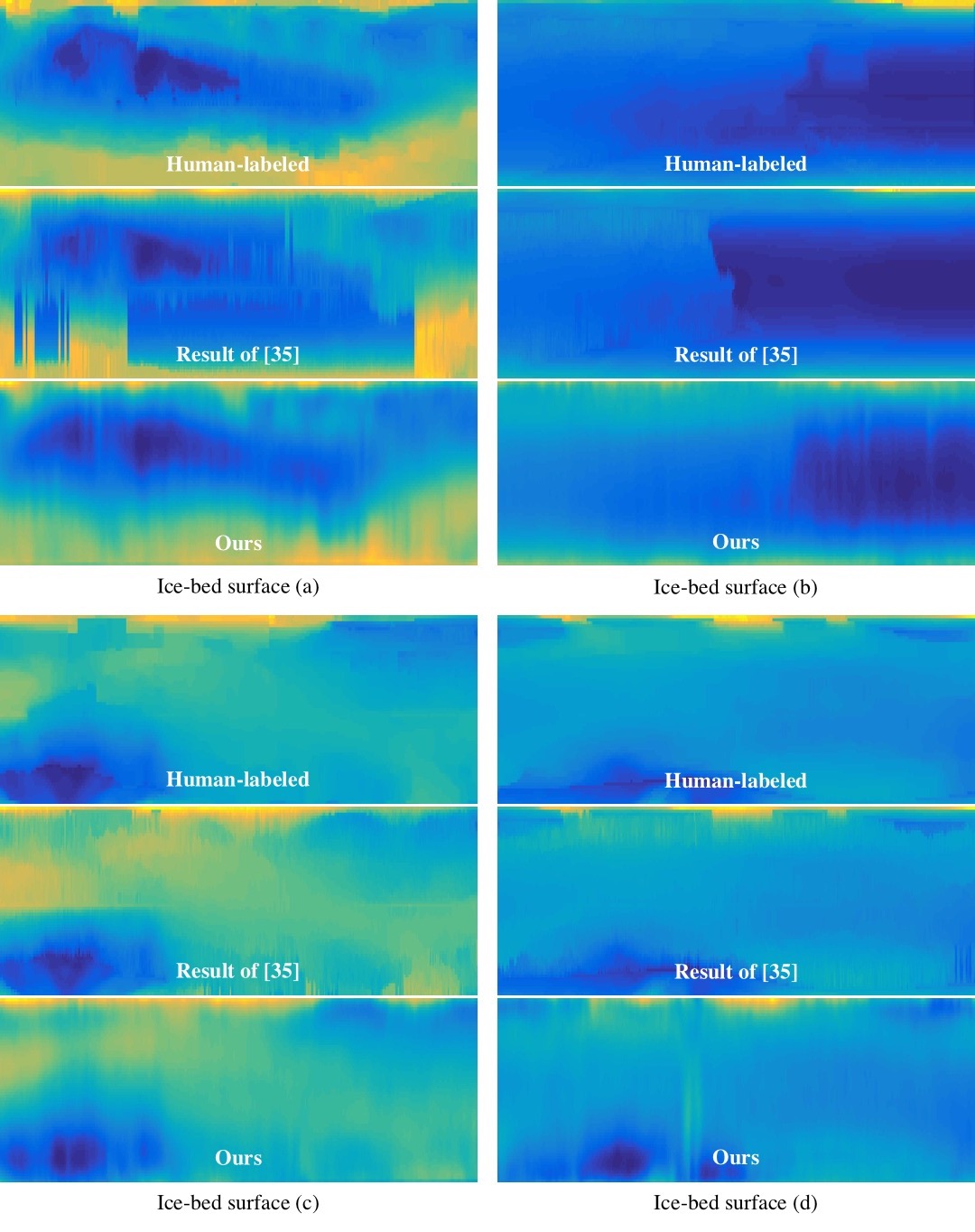}}
    \caption{Sample results of extracted ice-bed surfaces from a sequence of
        about 330 tomographic images.
 The x-axis corresponds to distance along
the flight path, the y axis is the width of the tomographic images ($W$),
and the color is the height dimension (max height is $H$), which also represents
the depth from the radar.
    }
    \label{fig:ice_bed}
    \vspace{-5pt}
\end{figure*}

\begin{table}[t]\centering
    \vspace{12pt}

    {\footnotesize{
        \ra{1.2}
        \begin{tabular}{@{}lccc@{}} \toprule
            & \textbf{Averaged Mean Error} (pixels) & \textbf{Time} (sec) \\ \midrule
            Xu et al. \cite{icesurface2017icip} & 11.9 & 306.0 \\
            Ours (C3D + RNN) & 10.6 & 51.6 \\
            \bottomrule
        \end{tabular}
}}
\vspace{-4pt}
    \caption{Performance evaluation compared to the state of the art. The
        accuracy of our approach is computed on the average of the ice-air and
        ice-bed surfaces {and the accuracy of~\cite{icesurface2017icip} is
        computed only on the ice-bed surfaces.} The running time is measured by
        processing a sequence of 330 tomographic images.
\vspace{10pt}
    }
    \label{tab:speed}
\end{table}

\begin{table}[t]\centering
    {\footnotesize{
        \ra{1.2}
        \begin{tabular}{@{}lccc@{}} \toprule
            & \multicolumn{2}{c}{Mean Error} \\
            \cmidrule{2-3}
            & \textbf{Ice-air surface} & \textbf{Ice-bed surface} \\ \midrule
            Crandall \cite{crandall2012layer} & --- & 101.6 \\
            Lee \cite{lee2014estimating} & --- & 35.6 \\
            Xu et al. (w/o ice mask) \cite{icesurface2017icip} & --- & 30.7 \\
            Xu et al. \cite{icesurface2017icip} & --- & 11.9 \\ \midrule
            Ours (RNN)       & 10.1 & 21.4 \\
            Ours (C2D)       & 8.8 & 15.2 \\
            Ours (C3D)       & 9.4 & 13.9 \\
            Ours (C2D + RNN) & 8.4 & 14.3 \\
            Ours (C3D + RNN) & 8.1 & 13.1 \\
            \bottomrule
        \end{tabular}
}}
\vspace{-4pt}
    \caption{Error in terms of the mean absolute column-wise difference compared
    to ground truth, in pixels.}
    \label{tab:accuracy}

\end{table}

\subsection{Dataset}
We use a dataset of the basal topography of the Canadian
Arctic Archipelago (CAA) ice sheets, collected by the Multichannel Coherent Radar
Depth Sounder (MCoRDS) instrument \cite{rodriguez2014advanced}. It contains a
total of 8 tomographic sequences, each with over 3,300 radar images corresponding
to about 50km of flight data per sequence. 
For training and testing, we also
have ground truth that identifies the positions of two layers of
interest (the ice-air and ice-bed, i.e., $K=2$). Several examples of these
tomographic images and their annotations are shown in Figure~\ref{fig:slices}.

To evaluate our model, we split the data into training and testing sets (60\%
as training images, 40\% as testing images) and learn the model parameters from
the training images. More formally, we wish to detect the ice-air and ice-bed
layers in each image, then reconstruct their corresponding 3D surfaces from a
sequence of tomographic slices. 
We assume the tomographic sequence has size  $C \times D \times H
\times W$, where $C$ denotes the number of image channels (which is $1$ for our
data), $D$ is the number of slices in the sequence, and $W$ and $H$ are the dimensions of each slice. We also parameterize the output surfaces as sequences, $S^k = \{S^k_1,
S^k_2, \cdots, S^k_D\}$, and $S^k_{d} = \{s^k_{d,1}, s^k_{d,2}, \cdots,
s^k_{d,W}\}$, where $s^k_{d,w}$ indicates the row coordinate of the surface
position for column $w$ of slice $d$, and $s^k_{d,w} \in [1, H]$ since the
boundary can occur anywhere within a column. In our case, $k \in \{0, 1\}$ 
represents the ice-air and ice-bed surfaces, respectively. 

\vspace{6pt}
\noindent
\textbf{Normalization.}
Since images from different sequences have different sizes (from $824 \times
64$ pixels to $2000 \times 64$ pixels), we resize all input images to $64
\times 64$ by using bicubic interpolation. For each image, we
also normalize their pixel values to the interval $[-1, 1]$ and subtract the mean
value computed from the training images. Further, since the coordinates of the
ground truth labels $G^k_d = \{g^k_{d,1}, g^k_{d,2}, \cdots, g^k_{d,W}\}$ in
each image $I_d$ are in absolute coordinates, we follow \cite{toshev2014deeppose}
to normalize them to relative positions in each image. Formally, each ground
truth label is normalized as,
\begin{equation}
N(g^k_{d,w}) = 2 (g^k_{d,w} - H/2) / H,
\end{equation}
and we predict the absolute image coordinates $s^k_{d,w}$ as,
\begin{equation}
s^k_{d,w} = N^{-1}(\textrm{M}_{\theta}(I_d)),
\end{equation}
where $\textrm{M}_{\theta}$ denotes our model with learnable parameters
$\theta$.

\subsection{Implementation Details}
We use PyTorch \cite{pytorch}  to implement our model, and do the
training and all experiments on a system with Pascal Nvidia Titan X
graphics cards. Each tomographic sequence is divided into 10 sub-sequences on
average, and we randomly choose $60\%$ of them as training data and the
remaining $40\%$ for evaluation. We repeat this training process (each time
from scratch) three times and report the average statistics for evaluation.

For C3D training, we use the Adam \cite{kingma2014adam} optimizer to learn the
network parameters with batch size of 128, each containing 5 consecutive radar
images. The training process is stopped after 20 epochs, starting with a
learning rate of $10^{-4}$ and reducing it in half every 5 epochs. The RNN training
is applied with the same update rule and batch size, but uses learning rate
$10^{-3}$ multiplied by $0.1$ every 10 epochs.

\subsection{Evaluation}
We evaluate our model on estimating the ice-air and ice-bed surfaces from
tomographic sequences of noisy radar images. We run inference on the testing
sub-sequences and calculate the pixel-level errors with respect to the
human-labeled ground truth. We report the results with two summary statistics:
mean deviation and running time. As shown in Table~\ref{tab:speed}, the
mean error averaged across the two different surfaces is about 10.6 pixels (where the
mean ice-air surface error is 8.1 pixels and mean ice-bed surface error is 13.1 pixels), and the
running time of processing a topographic sequence with 330 images is about 51.6
seconds. Figure~\ref{fig:ice_air} and~\ref{fig:ice_bed} show some example
results of the ice-air and ice-bed surfaces, respectively.

To give some context, we compare our results to previous state of the
art techniques as baselines, and results are presented in
Table~\ref{tab:accuracy}. Our first two baselines are Crandall et
al.~\cite{crandall2012layer}, which detects the ice-air and ice-bed
layers by incorporating a template model with vertical profile and a
smoothness prior into a Hidden Markov Model, and Lee et
al.~\cite{lee2014estimating}, who use Markov-Chain Monte Carlo (MCMC)
to sample from the joint distribution over all possible layers
conditioned on radar images.  These techniques were designed for 2D
echogram segmentation and do not include cross-slice constraints, so
they perform poorly on this problem.  Xu et
al.~\cite{icesurface2017icip} does use
information between adjacent images and achieves slightly better
results than our technique (11.9 vs 13.1 mean pixel error), but 
that technique also uses more information. In particular, 
they incorporate additional non-visual metadata
from external sources, such as the ``ice mask'' which gives prior weak information about anticipated
ice thickness (e.g., derived from satellite maps or other prior data). When we removed the ice mask cue from their technique 
to make the comparison fair, our technique beat theirs by a significant margin (13.1 vs 30.7 mean pixel error).
Our approach has two additional advantages:
(1) it is able to jointly estimate both the ice-air and ice-bed surfaces
simultaneously, so it can incorporate constraints on the similarity of these
boundaries, and (2) it requires less than one minute to process an entire 
sequence of slices, instead of over 5 minutes for~\cite{icesurface2017icip}.

In addition to published methods, we also implemented several baselines to
evaluate each component of our deep architecture. Specifically, we implemented:
(a) a basic C2D network using the same architecture with the 3D network but
with 2D convolution and pooling operations; (b) the RNN network using the
extracted features from the C2D as the initial hidden state; (c) the C3D network
alone without the RNN; and (d) the RNN network alone without the C3D network. The
results of these baselines are also shown in Table~\ref{tab:accuracy}. 
The results show that all components of the model are important for achieving
good performance, and that the best accuracy is achieved by our full model.

%% file: conclusion.tex
\section{Conclusion}
We have presented an effective and efficient framework for reconstructing
smoothed and structured 3D surfaces from sequences of tomographic images using deep networks.
Our approach shows significant improvements over existing techniques: (1) 
extracts and reconstructs different material boundaries simultaneously;
(2) avoids the need for extra 
evidence from other instruments or human experts; and (3) improves the
feasibility of analyzing large-scale datasets by significantly decreasing  the
running time. 
%We also demonstrated the performance of our model by producing
%convincing ice-air and ice-bed surfaces results against the state-of-the-art.

%% file: acknowledgements.tex
\section{Acknowledgments}
This work was supported in part by the National Science Foundation (DIBBs
1443054, CAREER IIS-1253549), and used the Romeo cluster, supported by Indiana
University and NSF RaPyDLI 1439007. We acknowledge the use of data from CReSIS
with support from the University of Kansas and Operation IceBridge (NNX16AH54G).
CF was supported by a Paul Purdom Fellowship.
We thank Katherine Spoon, as well
as the anonymous reviewers, for helpful comments and suggestions on
our paper drafts.